\documentclass[letterpaper, 10 pt, conference]{ieeeconf}  

\let\labelindent\relax
\usepackage{enumitem}
\usepackage[linesnumbered,ruled,vlined]{algorithm2e}
\usepackage{mathrsfs}
\usepackage{amsmath, amssymb}
\usepackage{svg}
\usepackage{caption}
\usepackage{float} 
\usepackage{capt-of}
\usepackage{lipsum}

\newtheorem{Def}{Definition}

\usepackage{hyperref}
\usepackage{leftidx}

\usepackage{multirow}
\usepackage{booktabs} 
\usepackage{diagbox} 

\usepackage{graphicx}
\usepackage{lscape}

\newlist{mydescription}{description}{1}
\setlist[mydescription]{
  labelwidth=3em, 
  leftmargin=!, 
  labelsep=1em, 
  itemsep=0em, 
  align=parleft 
}

\usepackage{comment}
\usepackage{slashbox}
\usepackage{booktabs}
\usepackage{multirow}
\usepackage{makecell}
\usepackage{gensymb}

\IEEEoverridecommandlockouts                              

\title{\LARGE \bf
Simulating Automotive Radar with Lidar and Camera Inputs
}

\author{Peili Song$^{1}$, Dezhen Song$^{2}$, Yifan Yang$^{1}$, Enfan Lan$^{1}$, and Jingtai Liu$^{1*}$
\thanks{*This work is supported by the National Natural Science Foundation of China under Grant 62173189.}
\thanks{$^{1}$Peili Song, Yifan Yang, Enfan Lan and Jingtai Liu are with the Institute of Robotics and Automatic Information System, Nankai University, Tianjin 300350, China; Tianjin Key Laboratory of Intelligent Robotics, Tianjin 300350, China; and also with TBI center, Nankai University, Tianjin 300350, China (e-mail: \{peilisong,yangyifan,lef\}@mail.nankai.edu.cn; liujt@nankai.edu.cn).
}
\thanks{$^{2}$Dezhen Song is with the Department of Robotics, Mohamed Bin Zayed University of Artificial Intelligence (MBZUAI),
Abu Dhabi, United Arab Emirates (e-mail:dezhen.song@mbzuai.ac.ae).
        }
\thanks{*Corresponding author}
}

\begin{document}

\maketitle
\thispagestyle{empty}
\pagestyle{empty}

\begin{abstract}

Low-cost millimeter automotive radar has received more and more attention due to its ability to handle adverse weather and lighting conditions in autonomous driving. However, the lack of quality datasets hinders research and development. We report a new method that is able to simulate 4D millimeter wave radar signals including pitch, yaw, range, and Doppler velocity along with radar signal strength (RSS) using camera image,  light detection and ranging (lidar) point cloud, and ego-velocity.  The method is based on two new neural networks: 1) DIS-Net, which estimates the spatial distribution and number of radar signals, and 2) RSS-Net, which predicts the RSS of the signal based on appearance and geometric information. We have implemented and tested our method using open datasets from 3 different models of commercial automotive radar. The experimental results show that our method can successfully generate high-fidelity radar signals. Moreover, we have trained a popular object detection neural network with data augmented by our synthesized radar. The network outperforms the counterpart trained only on raw radar data, a promising result to facilitate future radar-based research and development.

\end{abstract}

\section{INTRODUCTION}\label{sec:intro}

Low-cost automotive radars work at millimeter wave (mmWave) length and are widely used in autonomous driving due to its effectiveness under adverse weather and lighting conditions. During foggy, stormy, and smoky days, it is the last resort that can ensure the safety of vehicles on the road. However, quality datasets are limited due to existing research that mainly focuses on the use of regular camera and light detection and ranging (lidar) modalities, which hinders the development of radar-based navigation algorithms.  To address these issues, we propose using existing camera and lidar datasets, which are widely available, to simulate radar signal strength (RSS) and 4D radar signals, including pitch, yaw, depth, and Doppler velocity without the need of the prior 3D scene model or knowledge about material type.

\begin{figure}[htbp!]
  \centering
  \includegraphics[width=\linewidth, viewport=160 50 785 470, clip=true]{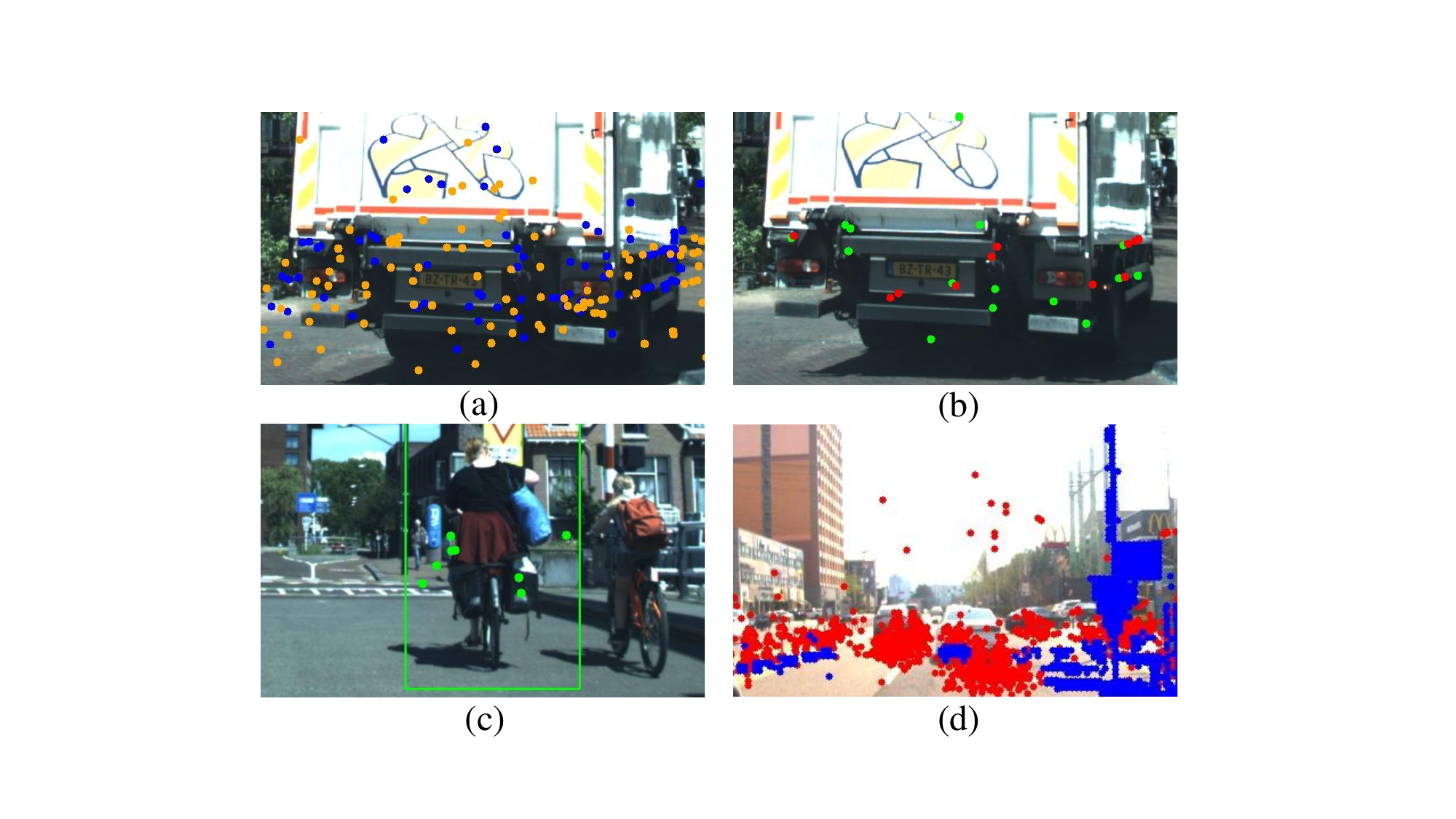}
  \caption{Properties of mmWave radar point cloud and the associated challenges of data synthesis: (a) Signal inconsistency: two radar point clouds  (brown and blue) from the same perspective do not have the same point positions. (b) Large depth variations: Green points are where radar points agrees lidar points in depth, red points otherwise. (c) Large point dispersion: radar signals reflected by a certain object may ``float'' around it. (d) Cross-modality sensitivity difference: radar (red) and lidar (blue) point cloud possess different reflection sensitivity.}
  \label{fig:properties_challenges}
\end{figure}

However, synthesizing radar points using images and lidar point clouds is non-trivial due to signal consistency, large depth variations, large point dispersion, and cross-modality sensitivity difference, as shown in Fig.~\ref{fig:properties_challenges}.  To address these issues, we employ learning-based approaches. First, we estimate radar signal distribution and number of reflections using a custom designed network that takes object appearance and radar ego-velocity as input.  The signal RSS values are generated for each point by fusing the multimodal information using a custom designed neural network which utilizes the information about local appearance and geometric shape to predict RSS without the need of precise 3D scene/material models or complex radiation pattern computation. Both networks can be easily trained using datasets with real radar signals. 
We have implemented the proposed algorithm and tested it using open datasets from 3 different commercial radar models. The ablation study results have shown that our design is successful. Signal distribution and quality are statistically indistinguishable from raw radar signals. Moreover, we have trained an object detection neural network with data augmented by our synthesized radar. The network outperforms the counterpart that only trained on raw radar data, a promising result for future radar-based research.

\section{RELATED WORKS}\label{sec:related}

Simulating radar datagram from camera images and lidar point clouds relates to the working principle of millimeter-wave radar, its applications in autonomous driving, and radar simulation techniques.

MmWave radar has radio signal wavelengths between 1.0mm and 10.0mm. It transmits and receives electromagnetic waves to obtain measures.  The RSS of the signal depends on the material, size, and structure of the reflection object. Typically, it is measured by radar cross section (RCS), which is equivalent to the cross-sectional area of a hypothetical sphere that is generated to measure the reflectivity of the target object \cite{harlow2024new}.  However, a low-cost automotive radar may/may not use RCS, but reports its own measured signal strength to characterize RSS in a nonstandard way. Therefore, in the paper, we will use the RSS as a broader measure instead of the RCS.  4D Mmwave radar plays a key role in autonomous driving systems due to its all-weather performance, high-precision range and velocity measurement, penetration capability, and resistance to interference \cite{fan20244d}. 4D mmwave radar data contains range, yaw, pitch, and Doppler velocity. Currently, the number of 4D radar datasets is very limited~ \cite{meyer2019automotive,palffy2022multi,zheng2022tj4dradset,choi2023msc}, which limits the development of radar-based navigation.

Due to the lack of available radar datasets, methods of simulating radar data have been explored. CARLA~\cite{dosovitskiy2017carla}, a well-known autonomous driving simulation platform, can only simulate radar data with very low fidelity. Vira \cite{schoffmann2021virtual} is based on Unity game engine using its built-in rendering pipeline.
4D radar data in \cite{cheng2023m}, a generic dataset, is generated by WaveFarer. 
To reduce the computational overhead of electromagnetic field simulations, a radar simulation framework based on a graphic simulation program, Blender, \cite{ouza2017simple} has been proposed.   There are also a few data-driven methods using generative models to generate radio-frequency (RF) data \cite{chen2023rf, deng2023midas} for certain human actions in static environments. These existing methods are either very limited in fidelity or have to rely on the computing radiation pattern that requires detailed knowledge such as 3D scene models and materials of particular targets, and frequency and antenna design of the simulated radar, all of which are often not available in practice. Our approach is the first of the kind to completely simulate 4D mmWave radar signals with radar signal strength information.

\section{Problem Formulation}\label{sec:formulation}

\begin{figure}[!h] 
\centering
\vspace*{-.15in}    \includegraphics[width=0.8\linewidth, viewport=35 110 940 460, clip=true]{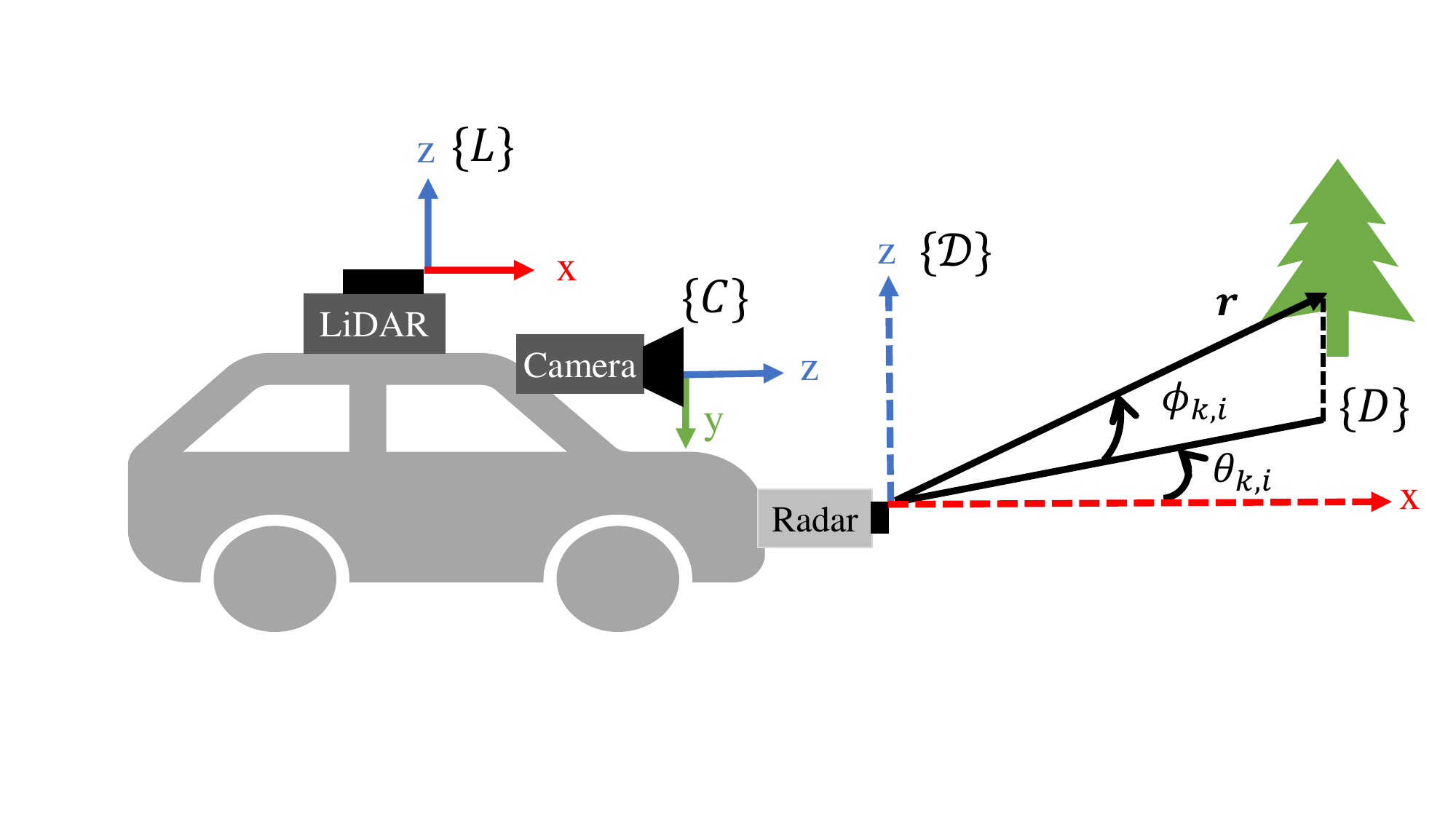}
  \captionof{figure}{Coordinate systems illustration.}
  \label{fig:coor}
\end{figure}

\noindent\textbf{Input, Output, and Coordinate Systems:}
At any given frame, we want to simulate a typical automotive radar using ego-velocity, a lidar point cloud, and a camera image. The coordinate systems are shown in Fig.~\ref{fig:coor}.

\vspace*{-.1in}
According to convention, a radar coordinate system $\{ D \}$ is a spherical coordinate system. The corresponding 3D Cartesian coordinate system  $\{\mathcal{D}\}$ is established with its origin coinciding with $\{D\}$, X-axis pointing forward, and Z-axis pointing vertically upwards. The lidar coordinate system $\{L\}$ and camera coordinate system $\{C\}$ are established as is shown in \ref{fig:coor}.  All 3D Cartesian systems are right-handed. 


The output radar datagram is defined as $\mathbf{D}:=\{\mathbf{s}_{i=1:n}\}$ where $i$ is the data index and $n$ is the maximum number of radar data points. Each reflected signal is in the spherical coordinates of $\{\mathbf{D}\}$ and contains:
\begin{equation}\label{eq:datagram_def}
    \mathbf{s}_{i}=\left[r_{i}, \theta_{i}, \phi_{i}, v_{i}, a_{\mbox{\tiny RSS},i}\right]^\mathsf{T},
\end{equation}
where $r_i$ is the range/depth, $v_i$ is the Doppler velocity, $\theta_{i}$ and $\phi_{i}$ are the azimuth and elevation (i.e. pitch and yaw) angles of the $i$-th signal, respectively, and $a_{\mbox{\tiny RSS},i}$ is the RSS value. Radars with such 5-element datagrams are often referred to as a 4D radar because it has first four elements in \eqref{eq:datagram_def} describing spatial and motion information. 

\vspace*{.05in}
\noindent\textbf{Assumption:}  Appearances of targets contain sufficient information of material and shape, which can determine the reflectivity of radar signals. 

Based on the assumption, with camera images providing appearance information such as shape and color, and lidar point clouds providing range/depth information, there are sufficient inputs for radar signal simulation.

\vspace*{.05in}\noindent\textbf{Nomenclature:} 
Let us define the important notations before introducing our problem. 
\begin{mydescription}

\item[$\mathbf{I}$]   the camera image ,
\item[$\mathbb{I}, \{I\}$]   the 2D image space $\mathbb{I} \subset \mathbb{R}^2$ and the associated coordinate $\{I\}$ with pixel coordinates $[u,v]^{\mathsf{T}}$ and origin located at top-left corner of the image, $u$-axis pointing leftward, and $v$-axis pointing downward,
\item[$\mathbf{l}_{j}$]        The $j$-th lidar point denoted as $\mathbf{l}_{j}=[x_{j},y_{j},z_{j}]^\mathsf{T} \in \mathbb{R}^3$, 
\item[$\mathbf{L}$]        lidar point cloud: $\mathbf{L}:=\{\mathbf{l}_{j=1:n_l}\}$, $n_l$ is the maximum number of lidar points,

\item[$_L^{\mathcal{D}}T$]   $4 \times 4$ homogeneous coordinate transformation matrix from $\{L\}$ to $\{\mathcal{D}\}$,
\item[$_C^{\mathcal{D}}T$]   $4 \times 4$ homogeneous coordinate  transformation matrix from $\{C\}$ to $\{\mathcal{D}\}$.

\end{mydescription}
Our problem is defined as follows.
\begin{Def}
 Given camera image $\mathbf{I}$, lidar point cloud $\mathbf{L}$, transformation matrices $_L^{\mathcal{D}}T$, $_C^{\mathcal{D}}T$, and radar's ego velocity $^\mathcal{D}\mathbf{v}_D$, generate corresponding radar datagram $\mathbf{D}$.
\end{Def}

It is worth noting that we are not attempting to generate raw radar signal, but to simulate a commercial low cost mmWave radar output which is usually much sparse after signal filtering, as shown in Fig.~\ref{fig:properties_challenges}. 

\section{Algorithms}\label{sec:algorithm}
\begin{figure}[htbp!]
  \centering
  \includegraphics[width=\linewidth, viewport=195 90 718 462, clip=true]{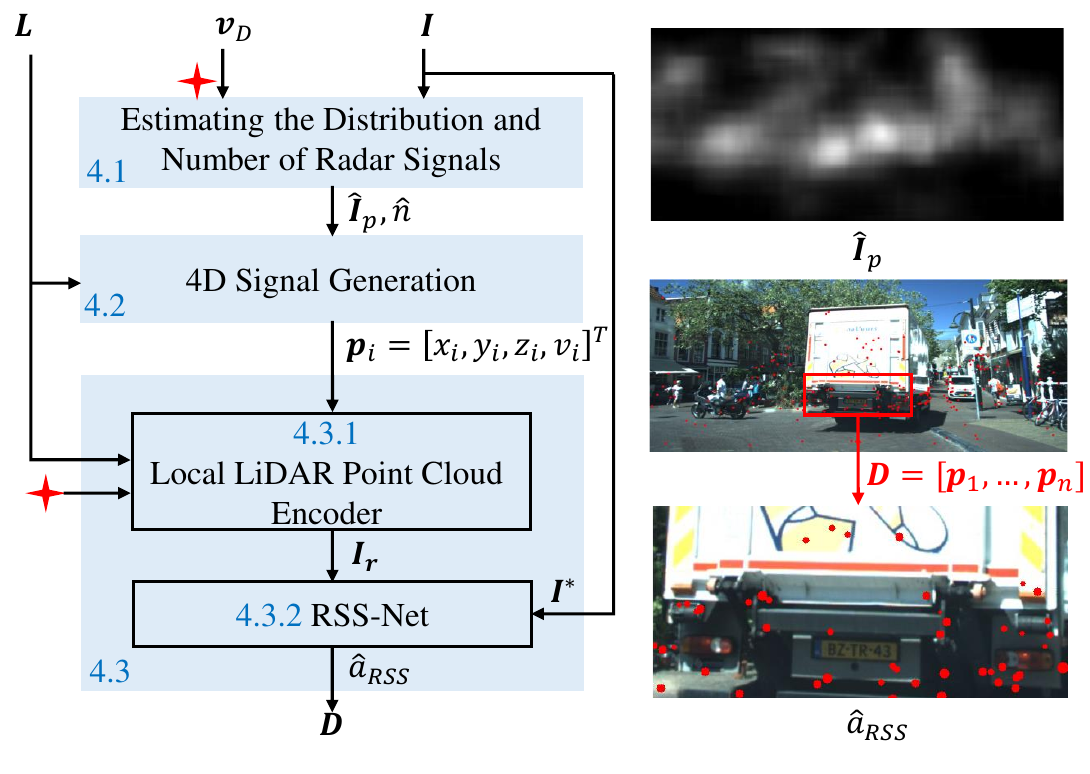}
  \vspace*{-.3in}\caption{Algorithm pipeline for generating simulated radar signals (inference stage). The algorithm takes lidar point cloud, camera image, and radar ego-velocty as input. The grayscale image represents the predicted radar signal distribution with high insensitivity representing high probability. The size of the red dot in the lower right image indicates the predicted RSS.}
  \label{fig:pipeline}
\end{figure}

Fig.~\ref{fig:pipeline} illustrates our algorithm pipeline which is consistent of three modules shaded in light blue.  Box 4.1 predicts radar signal distribution in the image frame and number of radar signals. Based on that, we generate 4D signals in Box 4.2. To complete the signal, RSS is generated in the last module. Let us explain them in sequence.

\subsection{Estimating the Distribution and Number of Radar Signals}\label{sec:distribution}
To simulate radar datagram, the first key issue is to determine where radar signals would be reflected and how many reflections would occur. As is shown in Fig.~\ref{fig:properties_challenges}~(a) and (c), a radar signal often disperse around the target due to mmWave radio signal refraction, reflection and scattering. This is very different from optical sensors at shorter wavelength. To address the issue, we design a Dis-Net to estimate the probability density function (PDF) and the total number of signals.





\begin{figure*}[htb!]
    \centering
     \includegraphics[width=4in,viewport= 5 150 955 515, clip]{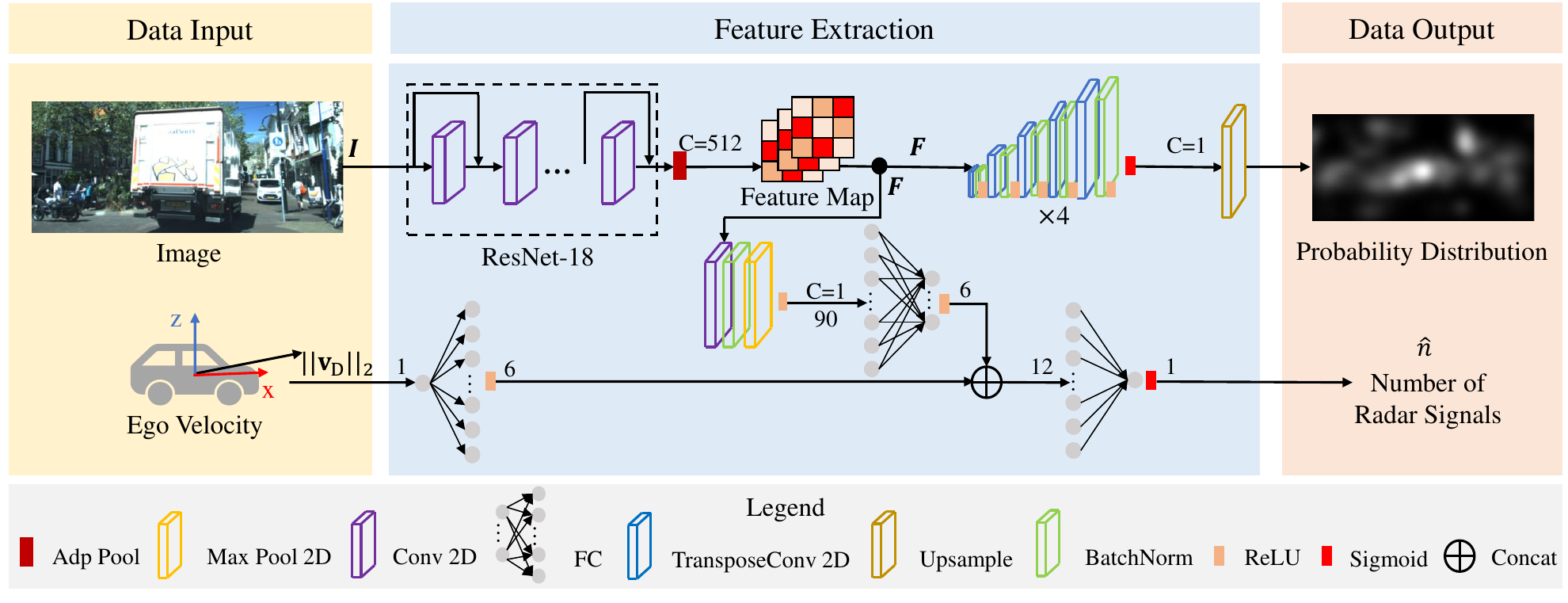}
    \caption{Overview of Dis-Net. The first branch generates a grayscale image as the probability density function of radar signals' distribution, and the second branch fits the number of signals. Avg Pool refers to average pooling, Adp Pool refers to adaptive pooling, Conv 2D refers to 2D convolutional layer, FC refers to fully connected layer (dense layer), TransposeConv 2D refers to 2D transpose convolutional layer, and Concat refers to concatenation. }
    \label{fig:DisNet}
\end{figure*}

\subsubsection{Dis-Net Architecture}

Fig.~\ref{fig:DisNet} illustrates the architecture of Dis-Net which has a two-branch structure: a distribution branch to estimate the probability distribution and a number branch to estimate the number of radar signals.

Let us first define the image input to be
\begin{equation}\label{eq:I_k}
    \mathbf{I} := \{(r,g, b)_{(u,v)}\bigl| u \in [1, u_{\mbox{\tiny max}}]\cap \mathbb{N}, v \in [1, v_{\mbox{\tiny max}}]\cap \mathbb{N} \},
\end{equation}  where $(r, g, b) \in [0, 255]\times [0, 255] \times [0, 255]$ are the red, green, and blue intensity values, respectively,  $u$ and $v$ are pixel coordinates, and $u_{\mbox{\tiny max}}$ and $v_{\mbox{\tiny max}}$ are maximum horizontal and vertical pixel indices, respectively.

To estimate radar signal distribution probability density in the image frame (represented as a grayscale image) $\mathbf{I}_{p}$, ResNet-18 \cite{he2016deep} is used as the backbone to encode $\mathbf{I}$ to be a high-dimensional feature map $\mathbf{F}$ containing appearance information such as color, texture, and semantic information.  $\mathbf{F}$ is
processed by a transpose convolutional network to restore the original image's size and reduce the number of channels to 1. An upsample module is used to ensure the output $\hat{\mathbf{I}}_{p}$ has the same size as $\mathbf{I}_{p}$.

The number branch predicts the number of reflections which requires the ego-velocity of the radar in addition to the camera image. Most objects has non-zero Doppler velocity as the radar moves, and the signals with significant Doppler velocity are likely to be retained after filtering \cite{song2014velocity} in radar signal processing algorithm. Therefore, only the magnitude of $\|^\mathcal{D}\mathbf{v}_D\|_2$ is utilized.
Feature map $\mathbf{F}$ and  $\|^\mathcal{D}\mathbf{v}_D\|_2$ are each encoded by a fully connected (FC) layer with the same output size. They are concatenated and processed by the last FC layer to estimate $\hat{n}$, the number of signals. 
At the end of both branches, the Sigmoid function is employed to map the output values to the range $(0,1)$.

\subsubsection{Dis-Net Training Scheme}\label{sec:PDF}
\begin{figure}[htbp]
  \centering
   \includegraphics[width=0.8\linewidth,viewport=75 98 900 440, clip]{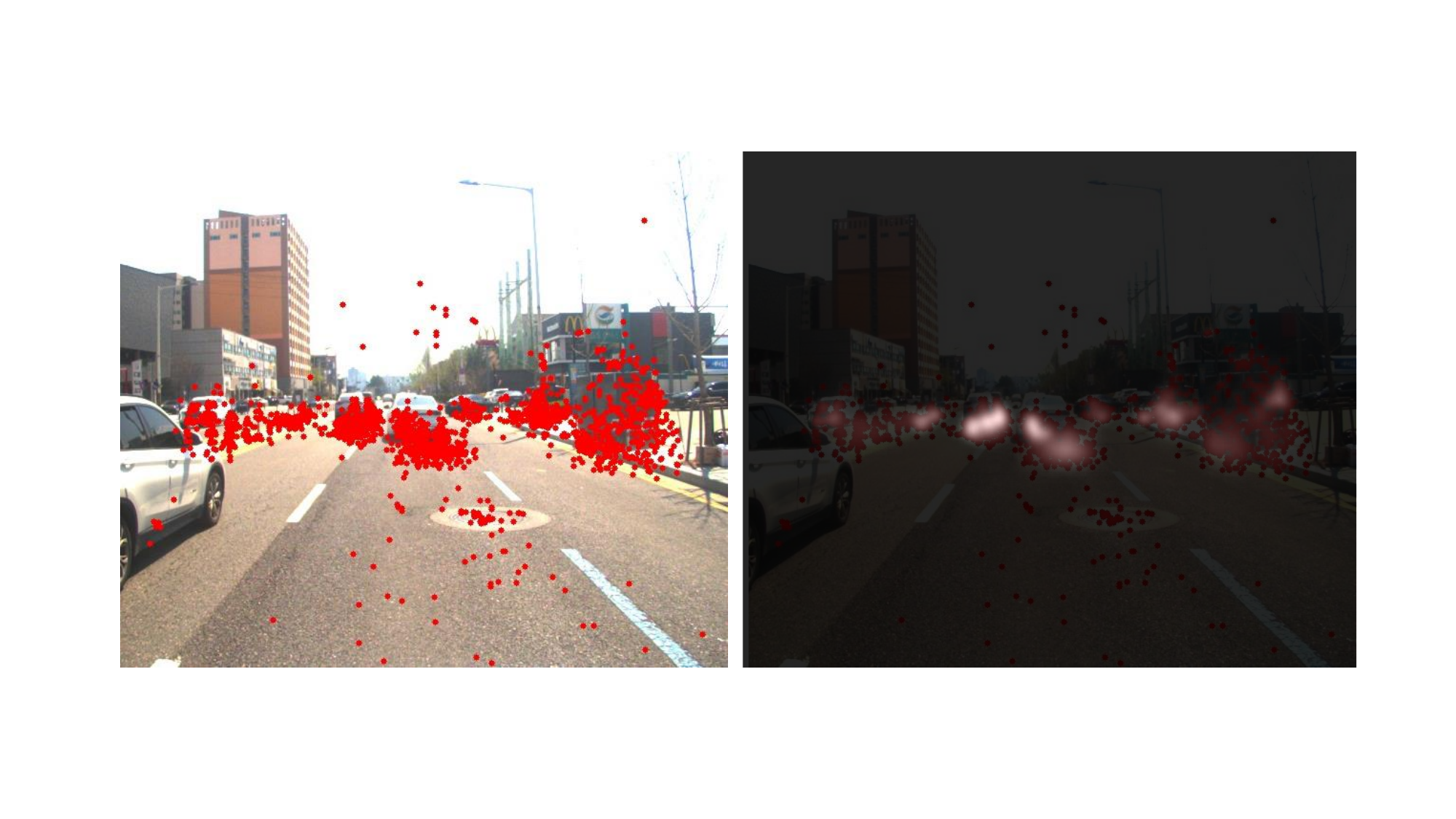}
  \caption{Left: radar signal (red dots) projection on image, and Right: the corresponding signal distribution function illustrated in the grayscale image (translucent).}
  \label{fig:grayscale}  
\end{figure}

To train Dis-Net, we employ real radar signals as ground truth. Because Dis-Net outputs the spatial signal distribution, it is necessary to generate the ground-truth radar signal distribution based on real radar readings. The first step is to project the radar signal from its polar coordinate to the image coordinate $\{I\}$. This is performed by first transferring  signal $\mathbf{s}_{i}$ in spherical coordinate to the collocated Cartesian coordinate,
\begin{equation}\label{eq:car2polar}^\mathcal{D}\mathbf{p}_{i}=
\begin{bmatrix}
x_{i} \\
y_{i} \\
z_{i}
\end{bmatrix}   
=
\begin{bmatrix}
r_{i} \cdot \cos(\phi_{i}) \cdot \cos(\theta_{i}) \\
r_{i} \cdot \cos(\phi_{i}) \cdot \sin(\theta_{i}) \\
r_{i} \cdot \sin(\phi_{i})
\end{bmatrix},
\end{equation}
where $^\mathcal{D}\mathbf{p}_{i}$ is the corresponding coordinate in $\{\mathcal{D}\}$, which is then converted to coordinate $\{ I\}$ using pinhole model \cite{sturm2021pinhole}

\begin{equation}\label{eq:pinhole}
\begin{bmatrix}
^I\mathbf{p}_{i} \\
1
\end{bmatrix}=\lambda K \left[_\mathcal{D}^CR \  _\mathcal{D}^C\mathbf{t} \right] \begin{bmatrix}
    {^\mathcal{D}\mathbf{p}_{i}} \\ 
    1
\end{bmatrix}),
\end{equation}
where $K$ is the intrinsic matrix of the camera, $(_\mathcal{D}^CR,\  _\mathcal{D}^C\mathbf{t})$ are the rotation matrix and the translation vector for the coordinate transform from $\{\mathcal{D}\}$ to $\{C\}$, respectively, 
and $\lambda$ is the scaling factor. The projection leads to radar datagram  $^I\mathbf{D}=\{^I\mathbf{p}_1,...,^I\mathbf{p}_n\}$, where $^I\mathbf{p}_i=[u_i,v_i]^{\mathsf{T}} \in \mathbb{I}$. The observation point set allows us to establish a posterior probability distribution for radar signals using a Gaussian mixture model. 
For a potential radar signal to be located at $^I\mathbf{x}$, each observation $^I\mathbf{p}_i$ determines the a 2D Gaussian distribution for $^I\mathbf{x}$ with the following probability density function (PDF), which is also known as 2D Bell function,
\begin{align}
\phi(&^I\mathbf{x};^I\mathbf{p}_i, \mathbf{\Sigma}) = \\
\nonumber &\frac{1}{2\pi}\sqrt{\mbox{det}(\mathbf{\Sigma})}\mbox{exp}\bigr(-\frac{1}{2}(^I\mathbf{x} -^I\mathbf{p}_i)^{\mathsf{T}} \mathbf{\Sigma}^{-1} (^I\mathbf{x} -^I\mathbf{p}_i)  \Bigl)\label{eq:gaussian-pdf},
\end{align}
where variance matrix $\mathbf{\Sigma}$ describes the spatial uncertainty of the signal due to the noisy nature of the radar. The value of $\mathbf{\Sigma}$ can be obtained through statistics of the 2D projection distribution of points from the annotated objects.

When there are many observations, the posterior radar signal distribution is modeled as a Gaussian mixture with the following aggregated PDF,
\begin{equation}\label{eq:mix_gaussian}
    p(^I\mathbf{x}) = \frac{1}{Z} \sum_{i=1}^{n} \phi(^I\mathbf{x};^I\mathbf{p}_i, \mathbf{\Sigma}),
\end{equation}
where and $Z$ is the normalization factor to ensure
\begin{equation}\label{eq:int_norm}
    \Sigma_{\leftidx{^I}{\mathbf{x}}{} \in \mathbb{I}} ~~ p(^I\mathbf{x})=1.
\end{equation}
The posterior signal distribution can be illustrated in image space $\mathbb{I}$ as $\mathbf{I}_{p}$ with each pixel (u,v) intensity  as $\mathbf{I}_{p}(u,v)=p([u,v]^\mathsf{T})\times 255$. Fig.~\ref{fig:grayscale} illustrates a sample $\mathbf{I}_{p}$. 

During training, Dis-Net predicts the radar signal distribution $\hat{\mathbf{I}}_{p}$ where symbol~ $\hat{}$~ is used to indicate the prediction of the network for the corresponding variable, which is a convention in the rest of the paper. On the other hand, $\mathbf{I}_{p}$ is used as the ground truth. To compare the similarity between two distributions, we employ Kullback–Leibler (KL) divergence
 as the loss function of the distribution branch:
\begin{equation}\label{eq:kl}
L_{\mbox{\footnotesize KL}}=\sum_{^I\mathbf{x} \in \mathbb{I}}\mathbf{I}_{p}(^I\mathbf{x})\cdot \mbox{log}(\frac{\hat{\mathbf{I}}_{p}(^I\mathbf{x})}{\mathbf{I}_{p}(^I\mathbf{x})}).
\end{equation}

The second branch is to estimate the number of signals where the ground truth $n$ is the length of real radar datagram $\mathbf{D}$. We propose to use the normalize relative difference as the loss $L_n$,
\begin{equation}\label{eq:loss_n}
    L_n=\frac{1}{m}(\sum_{i=1}^{m}\frac{\hat{n_i}-n_i}{n_i})^2,
\end{equation}
where $m$ is the number of frames, and $i$ is the frame index. To allow the network simultaneously predict the distribution and the number of signals, we combine total loss of Dis-Net as
\begin{equation}\label{eq:total_loss}
    L = L_{\mbox{\footnotesize KL}}+ 
    \alpha L_n,
\end{equation}
where $\alpha$ is a relative weighting  coefficient between the two branches.

\subsection{4D Signal Generation}\label{sec:4DGen}

In the inference phase, Dis-Net outputs the radar signal distribution $\hat{\mathbf{I}}_{p}$ and the signal number $\hat{n}$. Based on them, we can generate 4D radar signals through four steps: 1) signal generation, 2)  pitch and yaw generation, 3) range estimation, and 4) the Doppler velocity generation. 

\vspace{1mm}
\noindent\textbf{Signal Generation:}
Dis-Net's output $\hat{\mathbf{I}}_{p}$ and $\hat{n}$ allow us to draw $\hat{n}$ simulated radar signals. Recall that $\hat{\mathbf{I}}_{p}$ is the signal distribution function in grayscale image format, we can reconstruct the distribution function as follows,
\begin{equation}\label{eq:gray2prob}
    p(u,v)=\frac{\hat{\mathbf{I}}_{p}(u,v)}{\Sigma_{(x,y)\in \mathbb{I}}{\hat{\mathbf{I}}_{p}(x,y)}}.
\end{equation}
Based on the PDF, the task is to generate $\hat{n}$ signals in the image space $\mathbb{I}$ that can be obtained from the inverse transform sampling technique \cite{rendering2015physically} in $u$- and $v$- coordinates using its two-step random sampling process. Repeating it for $\hat{n}$ times, the process results in radar signals $\{[u_i, v_i]^{\mathsf{T}}_{i=1:\hat{n}}\}\subset \mathbb{I}$.

\vspace{1mm}
\noindent\textbf{Pitch and Yaw Coordinates Generation:}
For each generated radar signal $(u_i,v_i)$, the corresponding yaw and pitch angles in radar coordinate $\{D\}$ can be obtained as follows, 
\begin{equation}\label{eq:C2Polar}
    \begin{aligned}
    \theta_{i} &= \mbox{atan2} (y,x),\\
    \phi_{i} &= \mbox{arcsin} (z,\|[x, y, z]^{\mathsf{T}}\|_2),
    \end{aligned}
\end{equation}
where $[x, y, z]^{\mathsf{T}}=\leftidx{_C^\mathcal{D}}{R}{} K^{-1}[
    u_i, v_i, 1 ]^{\mathsf{T}}$ is the back projection of pinhole model in \eqref{eq:pinhole} and $\leftidx{_C^\mathcal{D}}{R}{} = \leftidx{^C_\mathcal{D}}{R}{^{-1}}$.



\vspace{1mm}
\noindent\textbf{Range Estimation:}
Range information can be estimated using the nearby lidar point range. For any point $^L\mathbf{p}_j \in \mathbb{R}^3$, it can be transformed to $\mathcal{D}$ as follows 
\begin{equation}\label{eq:lidar2radar}
    [^\mathcal{D}\mathbf{p}^{\mathsf{T}}, 1]^{\mathsf{T}} = \leftidx{_L^\mathcal{D}}T [^{L}\mathbf{p}^{\mathsf{T}}, 1]^{\mathsf{T}},
\end{equation}
and it can be further transformed to spherical coordinate $\{D\}$ using \eqref{eq:C2Polar}. We denote the transformed lidar point cloud by $^D\mathbf{L}=\{^D\mathbf{l}_{j=1:n_l}\}$, where $^D\mathbf{l}_j=[r_j,\theta_j,\phi_j]^\mathsf{T}$ and $n_l$ is the number of lidar points. Thus we can retrieve the nearby local point cloud through
\begin{equation}\label{eq:query_lidar}
    ^D\mathbf{L}^*=\{^D\mathbf{l}_j|^D\mathbf{l}_j \in ^D\mathbf{L},|\theta_i-\theta_j|\leq \delta_1, |\phi_i-\phi_j|\leq \delta_2\},
\end{equation}
where thresholds $\delta_1$ and $\delta_2$ are the simulated radar's horizontal and vertical angular resolution. Since lidar's angular resolution is higher than that of the radar, $^D\mathbf{L}^*$ must have $n_Q>0$ nearby points.
The range of radar signal $r_i$ is the average of that of the nearby points $^D\mathbf{L}^*$,
\begin{equation}\label{eq:depth_estimation}
    r_{i}=\frac{\sum_{j=1}^{n_Q} {r_j}}{n_Q}.
\end{equation}

\vspace{1mm}
\noindent\textbf{Doppler Velocity Generation:}
 Given the velocity of radar and target objects, Doppler velocity is the projection of the velocity vector to the unitary directional vector of the signal through vector dot product $< \cdot >$,
\begin{equation}\label{eq:doppler}
    v_{i}=\frac{<(^\mathcal{D}\mathbf{v}_O-^\mathcal{D}\mathbf{v}_D) \cdot \left(\leftidx{_C^\mathcal{D}}R K^{-1}[
    u_i, v_i, 1 ]^{\mathsf{T}}\right)>}{\|\leftidx{_C^\mathcal{D}}R K^{-1}[
    u_i, v_i, 1 ]^{\mathsf{T}}\|_2},
\end{equation}
where $^\mathcal{D}\mathbf{v}_O$ is the object's velocity which can be obtained by using the method in \cite{9830864} to distinguish signals reflected by background objects or dynamic objects.



\subsection{RSS Estimation}\label{sec:RSS}

\begin{figure}[htb!]
    \centering
     \includegraphics[width=3.2in,viewport= 0 110 920 515, clip]{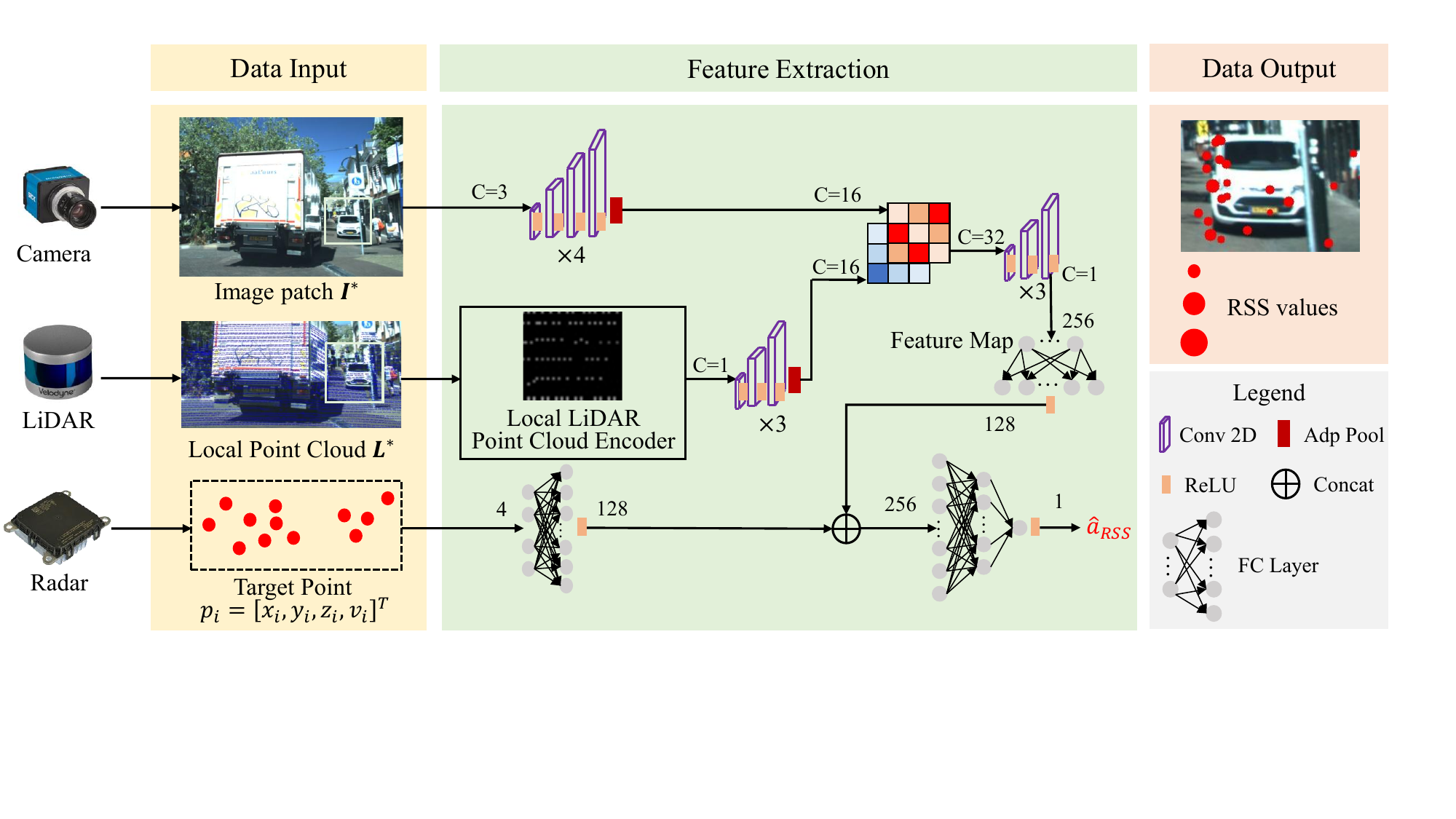}
    \caption{Overview of RSS-Net. The output is estimated RSS value, $\hat{a}_{\mbox{\tiny RSS}}$, and signals with larger RSS values are illustrated as larger red circles. }
    \label{fig:RSSNet}
\end{figure}

The next step is to generate the RSS values.  We design an RSS-Net, which is a multimodal neural network that integrates camera image appearance information, the local geometric feature of the lidar point cloud, and the generated 4D radar signals to predict RSS values $a_{\mbox{\tiny RSS}}$. Fig.~\ref{fig:RSSNet} illustrates the overall neural network design with the three types of input modalities where the 4D radar signals are used as anchor positions to identify the corresponding image patches and nearby lidar points. To facilitate the analysis, we use the 4D signals $^\mathcal{D}\mathbf{p}_i=[x_i, y_i, z_i, v_i]^\mathsf{T}$ in the Cartesian frame $\{\mathcal{D}\}$.

Given a radar signal $^\mathcal{D}\mathbf{p}_i$, the local image patch is first obtained to capture appearance information of nearby objects.
$^\mathcal{D}\mathbf{p}_i$ can be projected in image coordinate $[u_i,v_i]^\mathsf{T}$ using \eqref{eq:pinhole}. Its neighboring pixel patch is 
\begin{equation}\label{eq:I^*}
    \mathbf{I}^*:=\{(u,v)|u \in (u_i-r_c,u_i+r_c],v \in (v_i-r_c,v_i+r_c]\}, 
\end{equation}
where $r_c$ determines the neighboring range. 

\subsubsection{\textbf{Local Point Cloud Encoder}}
Lidar point clouds contains shape information and should be included as RSS-Net input. Again, since only the neighboring lidar points directly affect radar reflectivity, we only focus on those points. First, we convert lidar data to radar coordinate and denote the data as $^\mathcal{D}\mathbf{L}$ using \eqref{eq:lidar2radar}.
For each target radar signal $^\mathcal{D}\mathbf{p}_{i}$, local lidar point cloud is
\begin{equation}\label{eq:L^*}
    ^\mathcal{D}\mathbf{L}^*:=\{\mathbf{p}|\mathbf{p}\in\leftidx{^\mathcal{D}}{\mathbf{L}}{}, \| \mathbf{p}-\leftidx{^\mathcal{D}}{\mathbf{p}_{i}}{}\|_2 \leq r_l\}, 
\end{equation}
where $r_l$ is neighboring distance parameter.

To ensure the shape encoding of target objects is not affected by the position of $^\mathcal{D}\mathbf{p}_i$, the coordinate of any local point is translated and then projected
to pixel coordinates $(u_j, v_j)$ as follows,
\begin{equation}\label{eq:local_projection}
    u_{j}=\left\lfloor\frac{1}{2}(1-\frac{y_j-y_i}{r_l}) w\right\rfloor,
    v_{j}=\left\lfloor(1-\frac{z_j-z_i+r_l}{2r_l}) h\right\rfloor,
\end{equation}
where $^\mathcal{D}\mathbf{l}_j=[x_j,y_j,z_j]^\mathsf{T}$, $^\mathcal{D}\mathbf{p}_i=[x_i,y_i,z_i]^\mathsf{T}$, $w$ and $h$ are the width and height of the grayscale image, respectively, and $u_{j} \in [0,w)$, $v_{j} \in [0,h)$. The depth is encoded into grayscale $\mathbf{I}_r$ for all pixels as
\begin{equation}\label{eq:I_r}
    \mathbf{I}_r := \{{  r}_{(u_{j},v_{j})}, \forall (u_{j},v_{j})\},
\end{equation}
where
\begin{equation}\label{eq:grayscale_value}
    r = 
\begin{cases} 
127 + \lceil(\frac{\|^\mathcal{D}\mathbf{l}_j - ^\mathcal{D}\mathbf{p}_i\|_2}{2r_l}) \cdot 255 \rceil, &  \|^\mathcal{D}\mathbf{l}_j\|_2 \geq \|^\mathcal{D}\mathbf{p}_i\|_2 \\ 
127-\lfloor(\frac{\|^\mathcal{D}\mathbf{l}_j - ^\mathcal{D}\mathbf{p}_i\|_2}{2r_l})\cdot 255\rfloor, & \|^\mathcal{D}\mathbf{l}_j\|_2 < \|^\mathcal{D}\mathbf{p}_i\|_2
\end{cases},
\end{equation}
is the normalized depth in grayscale. If multiple points are mapped to the same pixel, the corresponding grayscale value is averaged. $\mathbf{I}_r$ is also called the range image \cite{stiene2006contour,steder2009robust} which is used as the input of the RSS-Net.

\subsubsection{Network Architecture and Training Scheme}

Fig.~\ref{fig:RSSNet} illustrates the overall network structure of the RSS-Net. The image patches $\mathbf{I}^*$ are $2r_c \times 2r_c \times 3$ in input dimension.  For each corresponding local lidar point cloud, it is a $w \times h \times 1$  range image $\mathbf{I}_{r}$. The positions of both are determined by the corresponding radar point $^\mathcal{D}\mathbf{p}_{i}=[x_{i},y_{i},z_{i},v_{i}]^\mathsf{T}$. It is worth noting that the Doppler velocity of the target point is also used as input because moving targets also tend to trigger stronger radar signals due to its filtering algorithm.

Due to the limited input size for RSS prediction, we design a lightweight neural network instead of utilizing the ResNet backbone. As shown in Fig.~\ref{fig:RSSNet},  image patches and local lidar point clouds are processed separately by two convolutional networks whose outputs are the same in dimension. The outputs are then concatenated to form a 32-channel feature map. It is further processed by a convolutional network to reduce the channel number to 1, so it is flattened and input to a FC layer. The output tensor contains both appearance and geometry information. It is concatenated with the feature of the embedded target point, and the concatenated tensor is finally processed by a multilayer perceptron (MLP) to generate the estimated RSS $\hat{a}_{\mbox{\tiny RSS}}$.

In the training stage, we employ signals from the real radar datagram that include ground truth signal positions, Doppler velocity, and RSS value $a_{\mbox{\tiny RSS}}$.
Similar to \eqref{eq:loss_n}, the following loss function is the normalized RSS difference and used to measure the performance of the network and to optimize parameters:
\begin{equation}\label{eq:loss_RSS}
    L_{a} = \frac{1}{m} \sum_{i=1}^{m} \Bigl(\frac{a_{\mbox{\tiny RSS},i} -\hat{a}_{\mbox{\tiny RSS},i}}{a_{\mbox{\tiny max}} - a_{\mbox{\tiny min}}} \Bigr)^2,
\end{equation}
where $m$ is the total number of signals, and $a_{\mbox{\tiny max}}$, $a_{\mbox{\tiny min}}$ are maximum and minimum  RSS values, respectively.

\begin{figure*}[htb!]
  \centering
   \includegraphics[width=\linewidth,viewport=38 168 886 380, clip] {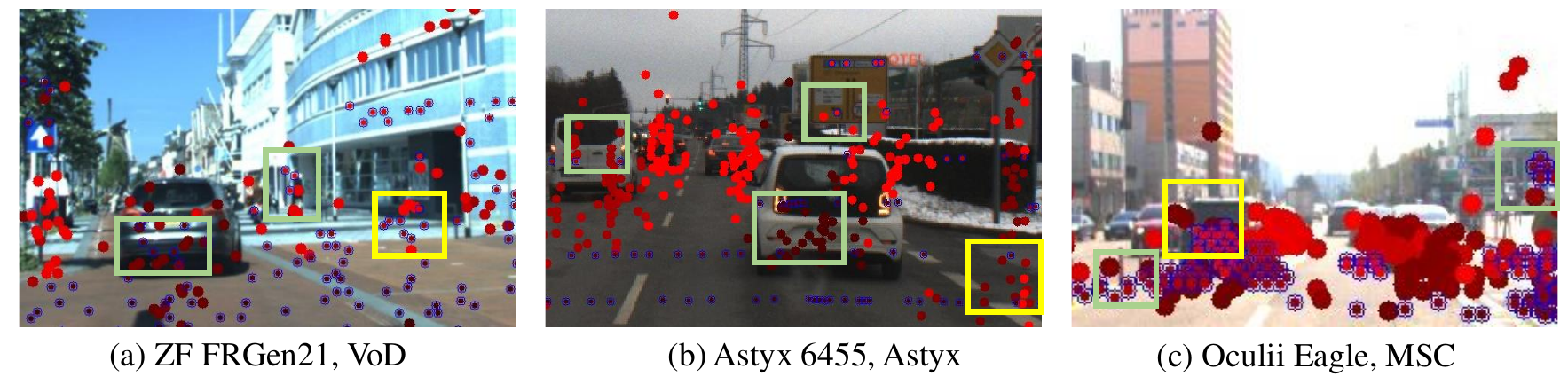}
   \vspace*{-.1in}
  \caption{Mmwave radar (big dots) and lidar (small dots with blue outer ring) point cloud projections in representative scenes on three datasets. The color of dots illustrates depth of each point, gradually changing from black (0 m) to red (50 m). Green and yellow bounding boxes represent good and bad depth consistency for mmwave radar and lidar, respectively. Calibration errors, synchronization errors, and differences in sensor properties cause inconsistencies. }
  \label{fig:datasets_proj} 
\end{figure*}

\section{EXPERIMENTS}\label{sec:experiment}

We have implemented the proposed Dis-Net and RSS-Net using PyTorch. The training and evaluation procedures are conducted with an NVIDIA RTX 3090 graphics card. Adam optimizer \cite{kingma2014adam} is used for both networks, and the learning rate is $1\times 10^{-4}$ for both networks. For RSS-Net, the shape of lidar range image is set to $128 \times 32$, which is determined by the ratio of horizontal and vertical angular resolutions of lidar.

\subsection{Dataset Illustration}\label{sec:ex_dataset}
\begin{table*}[htb!]
    \centering
    \footnotesize{
    \caption{Dataset information. \#Lines refer to the number of lidar scan lines. Avg \#Signal/F refers to Average number of signals per frame. \#Train (F) refers to number of the training frames. $\delta_1$ and $\delta_2$ are the horizontal and vertical angular resolutions of radar, respectively. Note that pitch and yaw angular resolutions of radar $\delta_1$ and $\delta_2$ for Astyx are estimated from the data due to lack of documentation.
   }
    
    \begin{tabular}{cccccccccc}
        \toprule
        Dataset & Radar type &$\delta_1$ & $\delta_2$ & Avg \#Signal/F & Min/Max \#Signal/F & \#Lines & Image Resolution & \#Train (F) & \#Test (F)\\
        \midrule
        VoD & ZF FRGen21 & $1.5\degree$ & $1.5\degree$ & 278.26 & 49/661 & 64 & $1936 \times 1216$  & 6065 & 2617 \\
        Astyx & Astyx 6455 & $1.5\degree$ & $1.5\degree$ & 969.08 & 183/3629 & 16 & $2048 \times 618$  & 382 & 164 \\
        MSC & Oculii Eagle & $1.0\degree$ & $1.0\degree$ &  1639.43 & 978/2462 & 128 & $720 \times 540$ & 744 & 320 \\
        \bottomrule
    \end{tabular}
    \label{tab:dataset_info}}
\end{table*}

\begin{figure}[htb!]
  \centering
   \includegraphics[width=\linewidth,  viewport=361 21 667 537, clip]{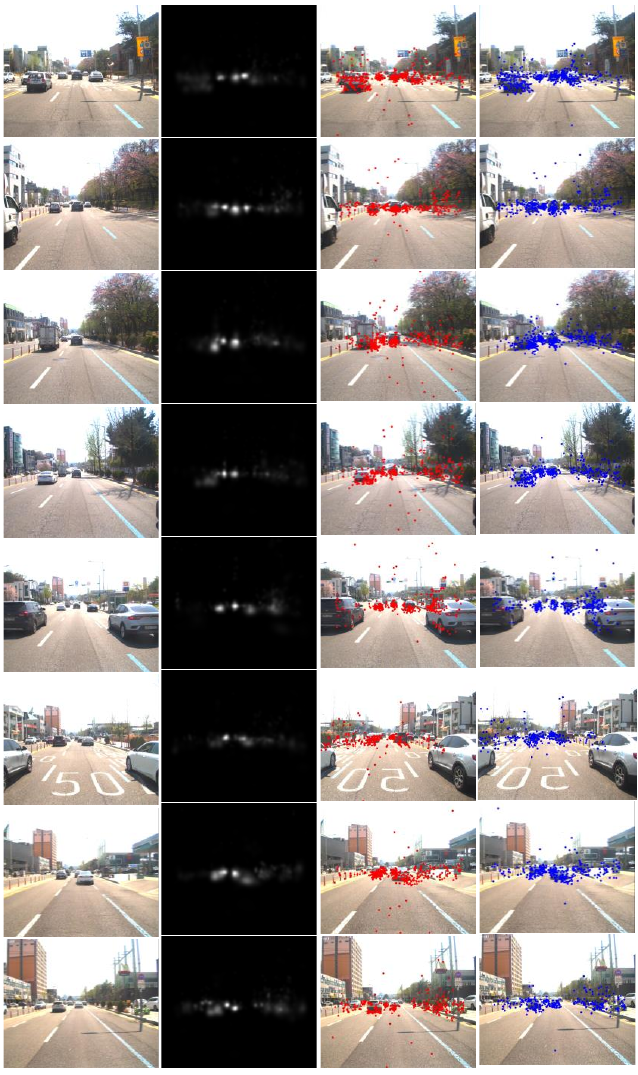}
  \caption{Performance visualization of Dis-Net. The four pictures in each row are raw images, predicted radar point probability distributions, and real (red) and synthesized (blue) radar point cloud projections. It is worth noting that closeup vehicles in the raw images may not register on radar and lidar modalities due to perspective difference.
}
  \label{fig:disnet_vis} 
\end{figure}

Experiments have been conducted on three open datasets with three different radar models: 1) View-of-Delft (VoD) dataset \cite{palffy2022multi}, 2) Astyx  2019 dataset \cite{meyer2019automotive}, and 3) MSC-Rad4R dataset \cite{choi2023msc}. The first two datasets provide rich annotations for object detection and tracking, and the third provides pose information for Simultaneous Localization and Mapping (SLAM). The detailed information is shown in Tab.~\ref{tab:dataset_info}.  VoD dataset contains 21 sequences, 13 of which were collected from residential areas, 7 from urban streets, and one mixed type.  Astyx  dataset includes only one sequence recorded on open roads. 
As for MSC-Rad4R (MSC) dataset, sequence URBAN{\_}D0 is utilized because it provides the most accurate calibration.



\begin{table}[htb!]	
	\centering
        \tiny
	\caption{Ablation study of Dis-Net.        Bold fonts indicate the best results.}
	\centering
	\begin{tabular}{cc|cc|cc|cc}
		\toprule[0.8pt]
		\multicolumn{2}{c}{} & \multicolumn{2}{c}{VoD} & \multicolumn{2}{c}{Astyx} & 
        \multicolumn{2}{c}{MSC}\\ \hline
		Image & Velo &  $L_{\footnotesize \mbox{\tiny KL}}$ & $L_n$   & $L_{\footnotesize \mbox{\tiny KL}}$ & $L_n$  & $L_{\footnotesize \mbox{\tiny KL}}$ & $L_n$  \\ \hline
		$\checkmark$ & $\times$  & \textbf{0.1538} & 0.0266  & 0.2359 & 0.4745  & 0.0539 & 0.3561  \\
		$\checkmark$ & $\checkmark$ & 0.1564 & 0.0266 & \textbf{0.2271} & \textbf{0.4499} &  \textbf{0.0524} & \textbf{0.3545} \\

		\toprule[0.8pt]
	\end{tabular}
	\label{tab:ablation_dis}	
	
\end{table}


\begin{table}[htb!]	
	\centering
    \footnotesize{
	\caption{Ablation study of RSS-Net. Bold fonts indicate the best results for each dataset. }
	\centering
	\begin{tabular}{ccc|c|c|c}
		\toprule[0.8pt]
		\multicolumn{3}{c}{} & \multicolumn{1}{c}{VoD} & \multicolumn{1}{c}{Astyx} & 
        \multicolumn{1}{c}{MSC}\\ \hline
		Cam & Li & RP & $L_a$  & $L_a$ & $L_a$  \\ \hline
        $\checkmark$ & $\times$ &$\times$ & $8.03\times10^{-3}$ & $1.53\times 10^{-2}$ & $1.50\times10^{-2}$\\
        $\times$ & $\checkmark$ &$\times$ & $5.89\times 10^{-3}$ & $1.51\times 10^{-2}$ & $2.77\times10^{-2}$\\
        $\times$ & $\times$ &$\checkmark$ & $5.31\times 10^{-3}$ & $1.53\times 10^{-2}$ & $2.47\times 10^{-7}$\\
        $\times$ & $\checkmark$ &$\checkmark$ & $4.14\times 10^{-3}$ & $1.47\times 10^{-2}$ & $7.27\times 10^{-8}$ \\
        $\checkmark$ & $\times$ &$\checkmark$ & $4.73\times 10^{-3}$ & $1.51\times 10^{-2}$ & $6.91\times 10^{-8}$  \\
        $\checkmark$ & $\checkmark$ &$\checkmark$ &$\mathbf{3.95 \times 10^{-3}}$& \textbf{$\mathbf{1.45\times 10^{-2}}$} & \textbf{$\mathbf{5.16\times 10^{-8}}$}  \\
       
		\toprule[0.8pt]
	\end{tabular}
	\label{tab:ablation_RSS}	
	}
\end{table}

The representative scenes and properties of the equipped radars/lidars are as shown in Fig.~\ref{fig:datasets_proj}. 
The radar signals in VoD dataset are less noisy, but the signal distribution is not concentrated on the target object. The radar signals in the Astyx dataset are more concentrated on moving vehicles and metal products, but they are extremely noisy, and the lidar point cloud is sparse. The radar signals in the MSC dataset accurately appeared near objects such as vehicles and road pillars, but the lidar did not successfully return readings for all vehicles. Due to the fact that depth inconsistency increases as depth itself increases, we only generate radar datagram within 50 meters at the intersection of the sensors' FoV.

To separate training and testing datasets for Dis-Net,  we randomly sample 30\% frames from sequences of different terrains/scene types for testing and leave the rest for training. The random sampling process ensures the distribution of different terrains/scene types in the training set and test set remains the same. The training set and test set of RSS-Net are obtained by randomly sampling radar signals from each frame of Dis-Net's training set and testing set. The numbers of sampled signals from VoD, Astyx, and MSC datasets are 50, 100, and 100 per frame, respectively, covering signals reflected from varies of ranges, angles, and objects. The parameter settings for RSS-Net is $r_c=50$ and $r_l=1$, which are tuned from 9 pairs of parameters drawn from sets $r_c \in \{25,50,125\}$ and $r_l \in \{0.5,1,2\}$.

\subsection{Ablation Study}\label{sec:ex_ablation}
We conduct an ablation study to evaluate the impact of various input modalities on the performance of both Dis-Net and RSS-Net.

Tab.~\ref{tab:ablation_dis} shows the results of the Dis-Net ablation study, showing that utilizing both image and ego velocity leads to better performances except on VoD dataset. It is worth noting that VoD dataset is collected by a radar with very sparse signals in low-speed scenes which means the Doppler velocity feature is not significant, thus the use of ego velocity does not obtain better results.

Tab.~\ref{tab:ablation_RSS} shows the results of the ablation study for RSS-Net. It is clear that utilizing all three input modalities, i.e. camera (Cam), lidar (Li),  and target radar point (RP), produce the best results, especially on datasets with cluttered and noisy radar signals. 
Moreover, for low-fidelity radar data, such as VoD and Astyx dataset, the target point and the local lidar point cloud are among the more important modalities in predicting RSS, which is not surprising, because radar signal attenuates over distance and the point cloud also carries shape information. Model trained on MSC dataset is not sensitive to the input modalities, because it provides high fidelity radar signals with less noise, making the network converge significantly better than others. There is a big difference before and after adding RP. This is due to the low resolution of the images, and the sensitivity difference between lidar and radar (as shown in Fig.~\ref{fig:properties_challenges} (d)).



\begin{table*}[htb!]
\caption{Performance comparison of Models PP$_{S_1}$ and PP$_{S_1 \cup S_3}$ (using $x,y,z,v$ features) on Car (IoU=0.5) and Cyclist (IoU=0.25), following the same settings as \cite{palffy2022multi}. Green up arrow \textcolor{green}{\(\uparrow\)} represents improvement, red down arrow \textcolor{red}{\(\downarrow\)} represents the opposite. The evaluation metrics are consistent with the KITTI open dataset \cite{geiger2013vision}. }
\begin{footnotesize}
    \centering
\begin{tabular}{@{}lcccccccccccc@{}}
\toprule
\textbf{Model} & \multicolumn{4}{c}{\textbf{Car (IoU=0.5)}} & \multicolumn{4}{c}{\textbf{Cyclist (IoU=0.25)}} \\ \midrule
               & \textbf{bbox AP} & \textbf{bev AP} & \textbf{3D AP} & \textbf{AOS AP} & \textbf{bbox AP} & \textbf{bev AP} & \textbf{3D AP} & \textbf{AOS AP} \\ \midrule
\textbf{PP$_{S_1}$} & 52.18 & 54.54 & 24.48 & 52.14 & 18.69 & 16.45 & 11.73 & 18.46 \\
\textbf{PP$_{S_1 \cup S_3}$} & 52.20 \textcolor{green}{\(\uparrow\)} & 54.46 \textcolor{red}{\(\downarrow\)} & 24.53 \textcolor{green}{\(\uparrow\)} & 52.16 \textcolor{green}{\(\uparrow\)} & 26.81 \textcolor{green}{\(\uparrow\)} & 21.30 \textcolor{green}{\(\uparrow\)} & 14.87 \textcolor{green}{\(\uparrow\)} & 26.77 \textcolor{green}{\(\uparrow\)} \\
\textbf{PP$_{S_1 \cup S_3}$}(with noise) & 53.23\textcolor{green}{\(\uparrow\)} & 54.21\textcolor{red}{\(\downarrow\)} & 24.74\textcolor{green}{\(\uparrow\)} & 53.17\textcolor{green}{\(\uparrow\)} & 19.34\textcolor{green}{\(\uparrow\)} & 16.58\textcolor{green}{\(\uparrow\)} & 7.73\textcolor{red}{\(\downarrow\)} & 19.18\textcolor{green}{\(\uparrow\)} \\ \bottomrule
\end{tabular}

\label{tab:sim2real}
\end{footnotesize}
\end{table*}

\subsection{Dis-Net Result Visualization}
The result of KL-divergence loss $L_{\footnotesize \mbox{\tiny KL}}$ and number loss $L_n$ in Tab.~\ref{tab:ablation_dis} actually show that the predicted signal distribution and number are consistent with the ground truth. To better demonstrate this conclusion, visualization is performed on the MSC dataset.  Fig.~\ref{fig:disnet_vis} shows typical results of the simulated radar datagram when compared to its ground-truth counterpart. Only $20\%$ of points within FoV are shown in the third and fourth columns to avoid cluttering. All sets of images show that the synthesized radar signal distribution is highly consistent with that of the real radar. More visualization results are shown in the attached video.

\vspace{1mm}
\subsection{Test Synthesized Radar Signals with Object Detection Task}

Next we show the synthesized radar datagram can be used to train an existing radar-based network for object detection, i.e. distinguishing cyclists and cars on the street. 
We follow the setup of the VoD dataset \cite{palffy2022multi} to implement training and testing process using PointPillars (\textbf{PP}) \cite{lang2019pointpillars}, which is available at OpenPCDet \cite{openpcdet2020}. 

Tab.~\ref{tab:sim2real} shows the results of \textbf{PP} training on only real data and on both real and synthesized data.
\textbf{PP} is first trained and tested on subset $S_1$ and $S_2$ from real radar data. The result is compared with another training scheme, training on $S_1 \cup S_3$ and testing on $S_2$, where $S_3$ is from synthesized radar data. 
$S_1$ and $S_2$, with length of 3149 and 1350, are sampled from the training set of Dis-Net. $S_3$,  with length of 723, comes from the inference results of Dis-Net on its testing set.


The results demonstrate that \textbf{PP$_{S_1 \cup S_3}$} outperforms \textbf{PP$_{S_1}$} significantly in cyclist detection, while maintaining a slightly better performance in detecting cars. This is due to the depth inconsistency between synthesized/lidar data and real data has a greater impact on detecting objects with larger bounding boxes. In addition, to further simulate the noise caused by scattering or multipath effects, $5\%$ of uniform noise is added to the synthesized data. Results show that training on dataset augmented by synthesized data improves for most metrics.

\section{CONCLUSIONS}

To assist development for radar-based vehicle navigation, we reported a new algorithm that can simulate automotive radar signals using an image, a lidar point cloud, and ego-velocity. The method does not require a precise 3D scene model or material type information for radiation pattern simulation. It directly predicted radar signal distribution and number using Dis-Net, a neural network built on a ResNet-18 backbone. We also designed a multimodal RSS-Net to predict RSS, a signal strength measure. We have implemented and tested the proposed method with datasets from 3 different commercial radars. Our experimental results have shown that the design was successful and the simulated radar signals have shown very high fidelity when comparing to the ground truth. 

In the future, we will further improve the network design to better utilize the multimodal information to improve fidelity and test the development in Sim2Real settings. 

\section{ACKNOWLEDGEMENT}

The authors gratefully acknowledge the support from the National Natural Science Foundation of China. We would also like to thank Peike Song for her valuable suggestions on figure design.








{
    \small
    \bibliographystyle{IEEEtran}
    \bibliography{main}
}

\end{document}